\documentclass{article}

\PassOptionsToPackage{numbers,sort&compress}{natbib}
\usepackage[preprint]{neurips_2026}

\usepackage[utf8]{inputenc}
\usepackage[T1]{fontenc}
\usepackage{hyperref}
\usepackage{url}
\usepackage{booktabs}
\usepackage{amsfonts}
\usepackage{amsmath}
\usepackage{amssymb}
\usepackage{nicefrac}
\usepackage{microtype}
\usepackage{xcolor}
\usepackage{graphicx}
\usepackage{float}
\usepackage{multirow}
\usepackage{array}
\usepackage{bm}
\usepackage{enumitem}
\usepackage{titlesec}
\usepackage{fontawesome5}

\titlespacing*{\paragraph}{0pt}{0.5ex plus 0.2ex minus 0.1ex}{0.6em}

\newcommand{\methodlong}{World Tracing}
\newcommand{\method}{WT}

\newcommand{\methodnet}{WT-DiT}
\newcommand{\objmodel}{WT-O}
\newcommand{\scenemodel}{WT-S}
\newcommand{\dynmodel}{WT-D}

\newcommand{\R}{\mathbb{R}}
\newcommand{\E}{\mathbb{E}}
\DeclareMathOperator{\relu}{relu}

\usepackage[table]{xcolor}

\definecolor{rankone}{RGB}{210,245,218}   
\definecolor{ranktwo}{RGB}{219,235,255}   
\definecolor{rankthree}{RGB}{255,242,204} 

\newcommand{\best}[1]{\cellcolor{rankone}\textbf{#1}}
\newcommand{\second}[1]{\cellcolor{ranktwo}#1}
\newcommand{\third}[1]{\cellcolor{rankthree}#1}

\title{\methodlong{}: Generative Pixel-Aligned Geometry\\ Beyond the Visible}

\author{%
  Hao Zhang$^{1,2}$ \quad
  Mohamed El Banani$^1$ \quad
  Jen-Hao Cheng$^1$ \quad
  Paul Zhang$^1$ \\
  Yi Hua$^1$ \quad
  Ben Mildenhall$^1$ \quad
  Christoph Lassner$^1$ \quad
  Narendra Ahuja$^2$ \quad
  Gengshan Yang$^1$ \\
  \normalfont $^1$World Labs \quad $^2$University of Illinois Urbana-Champaign \\[3pt]
  \normalfont\small
  \faGlobe~\href{https://haoz19.github.io/world-tracing-page/}{\texttt{Project Page}} 
  \quad
  \faGithub~\href{https://github.com/haoz19/world-tracing}{\texttt{Code}} 
  \quad
  \raisebox{-0.2ex}{\includegraphics[height=0.95em]{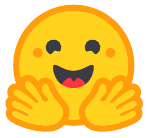}}~\href{https://huggingface.co/spaces/haoz19/world-tracing-demo}{\texttt{HuggingFace Demo}}
}

\begin{document}

\maketitle

\begin{figure}[H]
  \centering
  \includegraphics[width=\linewidth]{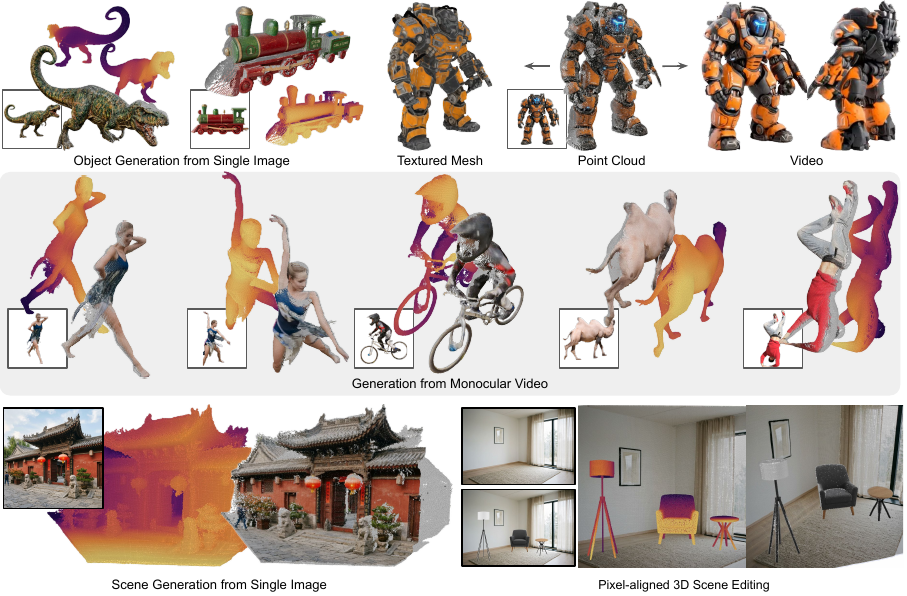}
  \vspace{-15pt}
  \caption{\textbf{\methodlong{}.} A pixel-aligned layered geometry representation that faithfully generates complete objects, scenes, and dynamic content from single images and monocular videos. 
  \textcolor{orange}{Colored points} are predicted visible surfaces; \textcolor{gray}{gray points} are predicted surfaces hidden from inputs. We visualize depth in magma colormap. This pixel-aligned representation enables several downstream applications: training-free pose-aware mesh generation, view synthesis, scene generation and editing.
  }
  \label{fig:teaser}
\end{figure}

\begin{abstract}
Image-to-3D methods often trade off faithfulness and completeness: depth estimators are anchored to input pixels but stop at the visible surface, while image-to-3D models generate complete shapes that are often misaligned to the input. We introduce \methodlong{}, a generative pixel-aligned geometry representation that predicts 3D points aligned with the observed pixels while completing geometry beyond the visible surface.
For each input pixel, \methodlong{} predicts an ordered stack of camera-space 3D points, where the first layer of the stack represents the visible surface and subsequent layers represent front-to-back intersections with occluded surfaces. 
We instantiate this representation with a world-tracing diffusion transformer (WT-DiT), which treats multiple geometry layers as separate denoising tokens coupled through factorized and global attention.
WT-DiT is trained with pixel-space flow matching and a mixed noise schedule that balances visible-surface reconstruction with occluded-geometry generation.
\methodlong{} achieves strong performance on visible-surface reconstruction and complete geometry generation across object, scene, and dynamic benchmarks, outperforming both depth predictors and image-to-3D generators. It also preserves 2D-to-3D correspondence, enabling text-driven 3D scene editing, geometry-conditioned novel-view video synthesis, and training-free integration with textured-mesh generators.
\end{abstract}

\section{Introduction}

Single-image 3D estimation has traditionally been explored through two main directions. The first faithfully reconstructs the 3D structure of the observed pixels, but is restricted to the visible surface~\citep{moge, moge2, dust3r, megasam, depthanything, depthanythingv2, marigold}. The second generates the full 3D object in a canonical frame, but does so at the cost of pixel alignment~\citep{trellis, sam3d, zero123, zero123plus, dreamgaussian, wonder3d}. As a result, neither paradigm provides downstream 3D pipelines what they need: faithful and complete geometry in the camera frame. The missing capability is \emph{faithful generation}: 3D generation that accurately reconstructs the visible surface and plausibly generates the invisible ones.


This representation choice also matters for data scaling. A pixel-aligned geometry representation can directly consume image-grid supervision, such as depth maps, rather than relying only on curated 3D assets.
Most foundation image-to-3D models~\citep{trellis} learn from datasets of artist-designed or scanned 3D assets that can be challenging to scale up and diversify. 
Moreover, their image backbones are often designed as a global latent encoder followed by a separate 3D generator. This design can discard fine-grained image evidence and weaken the pixel-level alignment needed for faithful reconstruction.
In contrast, a pixel-aligned design preserves local visual evidence throughout the geometry prediction process and allows the model to learn from image- and depth-supervised data at greater scale.

We introduce \methodlong{} (\method{}), a pixel-aligned multilayer geometry representation for this purpose. For each input pixel, \method{} predicts an ordered stack of L 3D points in camera space along the corresponding ray: the first layer is the visible surface, deeper layers complete the occluded geometry behind it. Faithful visible-surface reconstruction and generative completion are therefore not separate outputs, but successive layers of one tensor on the input pixel grid. The model predicts an image-grid pointmap and does not require camera intrinsics as input; when a pinhole intrinsics matrix is needed downstream, we fit a self-consistent $K$ in closed form from the predicted layer-$0$ geometry.

We instantiate this representation with \methodnet{}, a flow-matching diffusion transformer that treats multiple geometry layers as separate denoising tokens attending to each other. Since every layer lives on the image grid, the model inherits strong 2D visual priors from pre-trained image encoders, such as DINO~\cite{dinov2} or MoGe~\citep{moge}, rather than learning solely from rendered 3D assets. Unlike sparse multilayer formulations that require predicting per-layer validity masks~\citep{lari, ladi}, we use a depth-filling strategy: missing deeper-layer intersections are forward-filled in the target, while pixels outside the layer-$0$ alpha are noise-filled in the network input and ignored by the endpoint loss. This yields a single XYZ-only diffusion objective that trains all layers jointly. The same representation and core objective cover objects and scenes; the dynamic model only adds temporal attention.

We demonstrate three downstream uses of this representation in Sec.~\ref{sec:applications}: text-driven 3D scene editing, geometry-conditioned novel-view video synthesis, and training-free textured-mesh generation, all built without further per-task 3D training.

We evaluate \method{} across object, scene, and dynamic benchmarks with metrics that measure not only plausibility but also geometric consistency. \method{} surpasses monocular-depth predictors on visible-surface accuracy and image-to-3D generators on Chamfer distance to ground-truth geometry. Our contributions are:
\begin{enumerate}[leftmargin=*,itemsep=1pt,topsep=2pt,parsep=0pt]
\item A pixel-aligned multilayer geometry representation for faithful 3D generation, unifying high fidelity visible-surface estimation and occluded-geometry completion in one camera-space tensor.
\item \methodnet{}, a flow-matching diffusion transformer with efficient three-way factorized attention, trained using depth-filling objective that simplifies the prediction heads.
\item A comprehensive evaluation across objects, scenes, and videos showing that multilayer generation improves complete geometry while also improving visible-surface accuracy.
\item Downstream demonstrations: 3D scene editing, geometry-guided video synthesis, and training-free textured-mesh generation, showing \method{}'s effectiveness as a geometry prior for 3D pipelines.
\end{enumerate}

\section{Related Work}

\textbf{Monocular and multi-view 3D reconstruction.}
Pixel-aligned geometry has evolved from scale-ambiguous monocular depth \citep{eigen2014, midas, depthanything, depthanythingv2, unidepth} and depth-as-diffusion variants \citep{marigold, geowizard, lotus} to pointmap predictors that recover scale and intrinsics implicitly \citep{moge, moge2} and to multi-view reconstructors \citep{dust3r, mast3r, vggt, megasam}. All share one structural limit: a single 3D point per pixel, so geometry behind the first visible surface is absent. Layered depth images \citep{ldi, ldi_shade} and recent neural variants relax this by storing multiple surfaces per ray explicitly~\citep{peek, lari, ladi} or through an implicit  representation~\cite{saito2019pifu, saito2020pifuhd}; LaRI~\citep{lari} and DualPM~\citep{kaye2025dualpm} are our closest predecessors, which learn monocular layered depth \emph{regressor} for objects. \method{} extends this line by (i) covering objects, scenes, and dynamic clips with one architecture; (ii) replacing regression with flow-based diffusion models to represent multi-modal distributions of occluded surfaces; (iii) training at an order of magnitude larger scale; and (iv) inheriting image-based geometry priors from a frozen MoGe ViT-L encoder built on DINOv2~\citep{dinov2}.

\textbf{Image-to-3D generation.}
Image-to-3D pipelines fall into feed-forward generators \citep{trellis, lrm, instantmesh, sam3d, onetoplus}, multi-view-then-reconstruct approaches \citep{zero123, zero123plus, wonder3d, syncdreamer}, and SDS optimization \citep{dreamfusion, dreamgaussian, magic3d}, with diffusion backbones over radiance-field latents \citep{rgb2pc, ssd, 3dldm} or structured-latent / Gaussian spaces \citep{trellis, lgm}. They produce complete geometry, but typically in a canonical object frame, losing pixel alignment with the input. \method{} instead keeps the camera-aligned pixel grid as the native coordinate system. Recent hybrids \citep{reconviagen, lascomp} inject VGGT point clouds or features into TRELLIS; we compare against both as baselines for our training-free TRELLIS hybrid.

\textbf{Video-to-4D and feed-forward 4D.}
Optimization-based monocular 4D methods fit per-sequence NeRF / Gaussian / mesh / skeleton representations \citep{lasr, banmo, limr, ppr, pad3r, consistent4d, s3o, magicpose4d}, often relying on test-time optimization and auxiliary trackers or part templates. 
Feed-forward 4D predictors are faster, but they typically model dynamic geometry as single-surface point clouds, canonical meshes or Gaussians, structured spacetime latents, or trajectory fields~\citep{ss4d,gvf_diffusion,animateanymesh,actionmesh,spatialtrackerv2,traceanything}. In contrast, \dynmodel{} keeps the same pixel-aligned multilayer target as our static models and adds temporal attention to maintain coherence across frames.

\begin{figure}[t]
  \centering
  \includegraphics[width=\linewidth]{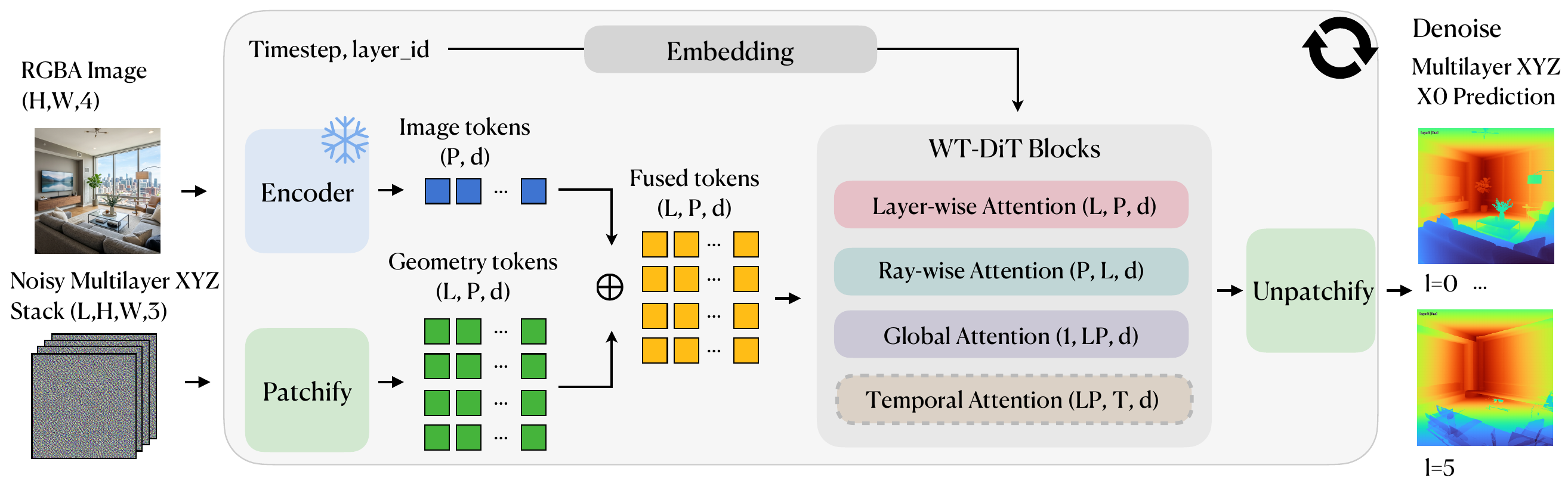}
  \vspace{-10pt}
  \caption{\textbf{\methodnet{} architecture.} A frozen MoGe encoder provides pixel-aligned image features, while noisy multilayer XYZ is patchified into geometry tokens. Pixel-aligned fusion combines image and geometry tokens before DiT decoder blocks with layer-wise, ray-wise, and global self-attention, plus temporal attention for \dynmodel{}. A linear patch projection maps each decoder token to the XYZ of its $14\!\times\!14$ patch, followed by unpatchification to the full multilayer image grid.}
  \label{fig:architecture}
\end{figure}

\section{Method}

We introduce an image-to-3D pipeline that produces $L$ layer pointmaps from a single image. A frozen MoGe ViT-L turns the input RGBA image into pixel-aligned image tokens, which are used to condition \methodnet{}, a flow-matching diffusion transformer that operates directly on the same image grid. Starting from Gaussian noise on a multilayer XYZ tensor, \methodnet{} integrates the flow ODE to produce a dense layered camera-space point stack: layer~$0$ recovers the visible surface, and deeper layers complete the occluded geometry along each pixel ray. Since every layer lives on the input pixel grid, the visual alignment with the input image and pixel-to-3D correspondences are preserved by construction.

\subsection{World Tracing Representation}
\label{sec:formulation}

Given an RGBA input $I{\in}\R^{H{\times}W{\times}4}$, an RGB image and a binary validity mask, \method{} predicts geometry as a pixel-aligned tensor of ordered 3D points
\begin{equation}
\mathbf{X}\;\in\;\R^{L\times H\times W\times 3},
\qquad
\mathbf{X}[\ell,\mathbf{u}] \;=\; \mathbf{x}_\ell(\mathbf{u}),
\end{equation}
where $\mathbf{x}_\ell(\mathbf{u})$ is the camera-space 3D point at the $\ell$-th front-to-back intersection of the ray through pixel $\mathbf{u}$. We do not require camera intrinsics as input: when needed, the parameters of a camera model can be recovered from the predicted layer-$0$ point cloud by fitting pixel-to-3D correspondences. The alpha channel of $I$ marks which pixels are valid: for the object model it isolates the foreground, and for the scene model it masks out infinite-depth pixels such as sky.

This representation turns faithful 3D generation into a layer-indexed multi-image generation problem. 
Layer~$0$ is the visible surface and is tightly constrained by the observed pixels. Deeper layers correspond to hidden surfaces, which are less directly observed and therefore require increasingly more conditional generation.
All predictions are made on the input pixel grid and expressed in the camera’s 3D coordinate system, so the output preserves the pixel-to-3D correspondences and camera pose information that canonical-frame generators typically discard.

\textbf{Dense targets without per-layer mask prediction.}
Previous depth peeling algorithms produce sparse pixels on later geometry layers -- many rays have fewer than $L$ true intersections. 
A direct multilayer formulation~\cite{lari,xray} would require predicting both depth/XYZ coordinates and a binary validity mask for each layer. This is difficult because deeper layers are valid on only a small fraction of rays, creating a severe class imbalance and encouraging mask collapse. It also couples coordinate regression with visibility classification, which we find can produce conflicting training signals.
Rather than ask the network to predict a separate visibility mask for each deeper layer, we make the supervision target dense by forward-filling empty intersections. If position $(\ell,\mathbf{u})$ has zero intersection, we fill it from the nearest earlier valid layer:
\begin{equation}
\mathbf{x}_\ell(\mathbf{u}) \;\leftarrow\; \mathbf{x}_{\ell'}(\mathbf{u}), \qquad \ell' = \max\{\, k < \ell : \mathbf{x}_k(\mathbf{u}) \text{ is valid}\,\}.
\label{eq:forwardfill}
\end{equation}
Let $\bar{\mathbf{X}}$ denote the resulting dense multilayer XYZ target. Every valid input ray therefore supervises all layers; the recovered shape is unchanged because filled entries are repeated points rather than new geometry. Modeling overlapping intersection points can express that some rays have already terminated by predicting deeper layers that collapse onto the front surface, avoiding the severe class imbalance of predicting sparse validity masks for deeper layers (Sec.~\ref{sec:exp_abl}).

\textbf{Scale normalization.}
Raw camera-space coordinates span very different ranges for objects and scenes, so we train in a normalized coordinate frame, $\tilde{\mathbf{X}}=\mathcal{N}(\bar{\mathbf{X}})$ using a invertible mapping chosen per regime. For objects, we use a dataset-level per-channel z-score normalization to standardize the target data to unit normal distribution; For scenes, whose depth can vary by orders of magnitude across samples, we keep them bounded by applying a per-sample median normalization, followed by a signed log transform. 
The same architecture predicts the normalized tensor in both regimes; only the coordinate transform changes. We provide the details in App.~\ref{app:scale_norm}.

\textbf{Flow-matching objective.}
We train \methodnet{} with flow matching \citep{rf, fm_lipman, ot_cfm, sd3} directly on the normalized coordinate signal rather than on a learned VAE latent~\citep{jit}. We set the clean endpoint to $\mathbf{x}_0=\tilde{\mathbf{X}}$, sample $\mathbf{x}_1\sim\mathcal{N}(\mathbf{0},\mathbf{I})$ and $t\sim p_{\mathrm{train}}(t)$, and form $\mathbf{x}_t=(1-t)\mathbf{x}_0+t\mathbf{x}_1$. Let $\mathbf{f}_I$ denote the projected MoGe feature grid. Architecturally, $\mathbf{f}_I$ is fused pixel-wise with the noisy geometry tokens before the decoder, rather than supplied through a separate conditioning branch. The network outputs an $\mathbf{x}_0$-parameterized prediction $\hat{\mathbf{x}}_0=F_\theta(\mathbf{x}_t^{\mathrm{net}},\,t,\,\mathbf{f}_I)$ trained with the endpoint loss
\begin{equation}
\mathcal{L}_{\mathrm{FM}} \;=\; \E_{\mathbf{x}_0,\mathbf{x}_1,t}\Bigl[\,\bigl\| A\odot\bigl(F_\theta(\mathbf{x}_t^{\mathrm{net}},\,t,\,\mathbf{f}_I) - \mathbf{x}_0\bigr) \bigr\|_2^2\Bigr],
\label{eq:fm}
\end{equation}
where $A$ is the alpha validity mask broadcast over layers and XYZ channels, so pixels outside the valid region are ignored by the loss. During training, the network input $\mathbf{x}_t^{\mathrm{net}}$ replaces the noisy geometry at invalid pixels with max Gaussian noise so the network learns to ignore them (App.~\ref{app:invalid_pixels}). At inference, the endpoint prediction induces the flow-ODE velocity $\mathbf{v}_\theta(\mathbf{x}_t,t,\mathbf{f}_I)=(\mathbf{x}_t-F_\theta(\mathbf{x}_t^{\mathrm{net}},t,\mathbf{f}_I))/t$, which we integrate from max noise towards $t{=}0$ using $20$ ODE steps.
The full loss adds a soft adjacent-layer monotonicity penalty $\mathcal{L}_{\mathrm{mono}}$ that pushes adjacent layers toward front-to-back order in the normalized coordinate space; because our normalization maps are monotone in depth, they preserve the layer ordering (App.~\ref{app:mono}). There is no per-layer mask, silhouette, or visibility classification head.

\subsection{\methodnet{} Architecture}
\label{sec:architecture}

\methodnet{} is a flow-matching diffusion transformer that maps an input image and a noisy multilayer point stack to an $x_0$ prediction on $\R^{L\times H\times W\times 3}$. Following recent feed-forward 3D transformers \citep{dust3r, vggt}, the design is largely standard: a frozen 2D foundation encoder produces pixel-level evidence, and a stack of pre-norm DiT blocks predicts the full layered pointmaps. We introduce two design-choices specific to  our multilayer representation: a three-way attention factorization (within-layer, along-ray, and global) and a FiLM layer embedding that breaks the layer permutation symmetry.

\textbf{Encoder and tokenization.}
Image evidence comes from a frozen MoGe ViT-L \citep{moge}; only a feature projection from its last blocks is trained, so \methodnet{} inherits MoGe's in-the-wild visual priors rather than learning them from rendered geometry. The noisy XYZ tensor is patchified on the same pixel grid (one geometry token per patch, per layer; $L{=}6$, discussed in App.~\ref{app:layers_choice}); at every $(\ell, \mathbf{u})$ the noisy geometry is concatenated with the repeated image feature and projected to the decoder width, so image evidence and geometry state stay in correspondence throughout the decoder without a separate cross-attention path. Hyperparameters and parameter counts are in App.~\ref{app:training}.

\textbf{Three-way attention factorization.}
Full attention over all $L\!\times\!P$ tokens is unstructured and costly, so the decoder cycles through three attention shapes: \textbf{layer-wise} attention $(B\!\cdot\!L,\,P,\,D)$, where each layer attends within itself as a 2D image with 2D RoPE over $(y,x)$; \textbf{ray-wise} attention $(B\!\cdot\!P,\,L,\,D)$, where tokens at the same pixel attend along the front-to-back layer axis, enforcing depth ordering and layer coherence at each ray; and \textbf{global} attention $(B,\,L\!\cdot\!P,\,D)$ that recovers object- or scene-level context at higher cost. Compared to the frame/global alternation used by VGGT \citep{vggt} for multi-view inputs, the explicit ray axis is what lets a single backbone produce coherent layered geometry from one image, and is a key reason that deeper layers do not drift away from the visible surface (Sec.~\ref{abl:mask}).

\textbf{Layer-aware conditioning.}
Each token must know which layer it represents and which diffusion time it is being denoised at. \textbf{Layer FiLM} maps a per-layer embedding $\mathbf{e}_\ell$ through an MLP to channel-wise $(\gamma_\ell,\beta_\ell)$ and applies feature-wise modulation $h \leftarrow \gamma_\ell\!\odot\!h+\beta_\ell$, which is enough to break the layer permutation symmetry without learnable additive position tokens. The diffusion time $t$ uses the standard AdaLN modulation \citep{sd3} shared across all tokens of the stack.

\textbf{Temporal attention for dynamic clips.}
For \dynmodel{}, we keep the static decoder unchanged and insert one temporal-attention block after each global-attention block, with 1D RoPE along the time axis and a small LayerScale init~\citep{layerscale} so that the single-frame \objmodel{} checkpoint reproduces the static behavior on $T{=}1$ inputs and gradually picks up temporal coupling during fine-tuning (App.~\ref{app:training}).

\subsection{Training and Model Variants}
\label{sec:impl}

\textbf{Model variants.}
We train three variants of the same model family with $L{=}6$ layers: \objmodel{} (static objects), \scenemodel{} (static scenes), and \dynmodel{} (dynamic objects). They share the representation, flow-matching loss, depth filling, frozen image encoder, and structured attention; they differ only in the scale normalization (z-score for object/dynamic, log-median for scenes), input resolution, and the temporal blocks added to \dynmodel{}. 
To stay robust to imperfect masks, we jitter the validity-mask boundary during training. Pixels whose support changes under this jitter are supervised with weak pseudo-targets obtained from the nearest valid rendered geometry after augmentation, and their losses are down-weighted.
Downstream applications use the self-consistent intrinsics estimated from the layer$-0$ pointmap, as described in Sec.~\ref{sec:applications} and App.~\ref{app:training}.

\textbf{Training noise curriculum.}
The visible layer and the occluded layers have different uncertainty profiles. Layer$-0$ is strongly constrained by the input image and behaves more like a reconstruction target, while deeper layers are only indirectly constrained and require conditional generation. A single diffusion-time distribution therefore under-serves one of these regimes.
We use a layer-aware curriculum: early in training, the visible layer is sampled from a plateaued logit-normal distribution and deeper layers from the standard logit-normal~\citep{sd3}; once all layers have stabilized, the whole stack switches to a shared timestep drawn from an equal mixture of the two schedules. This respects the different uncertainty profiles of visible and occluded geometry while preserving a single objective. Analytic densities of the three schedules are visualized in App.~\ref{app:noise_curriculum}.

\subsection{Data Pipeline}
\label{sec:rendering}
\textbf{Depth-peeling supervision.}
Training the multilayer representation requires per-ray, ordered ray-surface intersections, which are not provided by datasets curated for visible-surface depth or mesh generation. We therefore construct multilayer supervision by rasterizing curated 3D assets with depth peeling~\citep{everitt2001interactive,mammen1989transparency,nvdiffrast}. For each rendered image, the first \(L\) front-to-back intersections along every camera ray are unprojected into camera-space XYZ targets aligned with the tensor predicted by WT.

\textbf{Dense multilayer targets.}
Rays with fewer than \(L\) intersections are populated using the forward-filling rule in Sec.~\ref{sec:formulation}. This gives every valid input ray a dense \(L\)-layer XYZ target while preserving the recovered geometry, since filled entries repeat existing surface rather than introduce new geometry.

\textbf{Training corpora and augmentations.}
We build three corpora with the same representation: approximately 300K objects from public collections, scene frames from the public 3D-FRONT split plus a held-out internal scene set used only as an additional generalization probe, and approximately 16.8K dynamic clips. All renderings use randomized lighting, viewpoints, and intrinsics, and are combined with online geometric, photometric, and validity-mask augmentations whose camera-space targets are transformed jointly with the image. We provide asset sources, rendering settings, augmentation details, and per-layer occupancy statistics in Appendix~\ref{app:data}.

\begin{figure}[t]
  \centering
  \includegraphics[width=\linewidth]{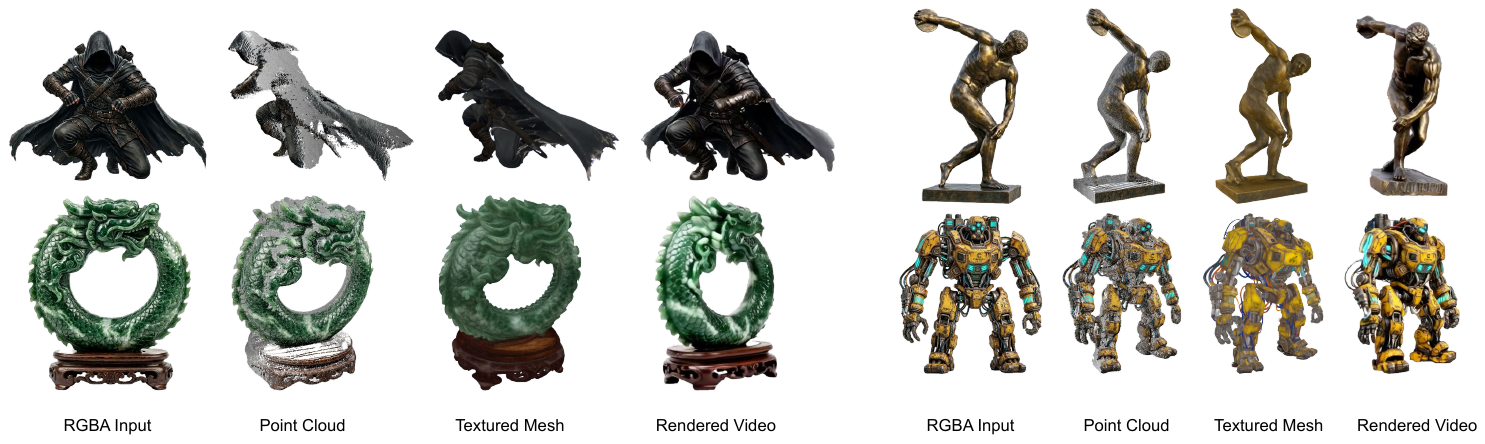}
  \vspace{-15pt}
  \caption{\textbf{Pixel-aligned geometry as a unified 3D interface.} \method{} generates complete multilayer geometry in the input camera frame while preserving pixel correspondences. This representation enables pose-aware structure for training-free textured-mesh pipelines such as TRELLIS-style decoders, and serves as geometry memory for novel-view video synthesis.}
  \label{fig:pc_mesh_video}
\end{figure}

\subsection{Mix-Training Across Multilayer and Single-Layer Supervision}
\label{sec:mixtrain}
Prior single-image depth models are usually tied to one supervision regime. Monocular depth and pointmap predictors expose only a single visible layer, while image-to-3D models such as TRELLIS requires full 3D geometry. This makes it difficult to train one model jointly from both full 3D asset geometry and single-layer RGBD captures.

Our representation supports both regimes without changing the architecture. Let \(b_{\mathrm{single}}\in\{0,1\}\) indicate whether a sample provides only visible-surface supervision or full multilayer supervision. We gate the loss-validity mask \(A\) from Eq.~\ref{eq:fm} along the layer axis:
\begin{equation}
A^{(\ell)} \;\leftarrow\; A^{(\ell)}\,\bigl(1 - b_{\mathrm{single}}\,\mathbb{1}[\ell{\ge}1]\bigr).
\label{eq:singlelayer_mask}
\end{equation}
Prior single-image depth models commit to one regime by construction: monocular depth predictors~\citep{moge2, pi3, depthanything3} expose a single layer and have no mechanism for multilayer supervision, while LaRI~\citep{lari} uses a depth-plus-per-layer-mask head trained only on synthetic multilayer renders and never sees real RGBD captures. The pixel-aligned representation of Sec.~\ref{sec:formulation}, the depth-filling target of Sec.~\ref{sec:formulation}, and the per-sample mask gate in Eq.~\ref{eq:singlelayer_mask} together let a single \method{} model train concurrently on multilayer 3D-asset renders and single-layer RGBD captures iteratively. The data mix used in practice is described in App.~\ref{app:data}, and the resulting lift on real-scene visible-surface benchmarks is reported in App.~\ref{app:real_scene_depth} (Table~\ref{tab:real_scene_depth}).

\section{Experiments}
\label{sec:experiments}

\begin{table}[t]
\centering
\setlength{\tabcolsep}{2.2pt}
\caption{\textbf{Object geometry.} Left: visible-surface depth on held-out objects after the same per-sample scale--shift alignment for all methods. Right: full object geometry and TRELLIS hybridization on a 100-sample object benchmark; stochastic diffusion/generative methods are reported as best-of-$8$ random seeds. DA3 denotes Depth Anything 3; Pi3X is the upgraded $\pi^3$ model; 3DGS denotes 3D Gaussians; PC denotes point clouds directly produced by \objmodel{}; \objmodel{}* denotes the 3D assets produced by combining \objmodel{} with TRELLIS.2's Stage 2 (detailed geometry generation) model. \small \textcolor{rankone}{\rule{0.9em}{0.9em}} Best \quad
\textcolor{ranktwo}{\rule{0.9em}{0.9em}} Second best \quad
\textcolor{rankthree}{\rule{0.9em}{0.9em}} Third best.}
\vspace{-2pt}
\label{tab:object_results}
\begin{minipage}[t]{0.49\linewidth}
\centering
\textbf{Visible surface depth}\\[-0.4ex]
\resizebox{1.00\linewidth}{!}{%
\begin{tabular}{lccccc}
\toprule
Method & MAE$\downarrow$ & RMSE$\downarrow$ & AbsRel$\downarrow$ & $\delta{<}1.25\uparrow$ & $\delta{<}1.25^2\uparrow$ \\
\midrule
DA3~\citep{depthanything3} & 0.0703 & 0.0920 & 0.0384 & 0.9973 & 0.9998 \\
LaRI~\citep{lari} & 0.0366 & 0.0506 & 0.0198 & 0.9992 & \best{0.9999} \\
Pi3X~\citep{pi3} & 0.0317 & 0.0440 & 0.0172 & 0.9994 & \best{0.9999} \\
MoGe-2~\citep{moge2} & \third{0.0261} & \second{0.0368} & \third{0.0141} & \second{0.9995} & \best{0.9999} \\
VGGT~\citep{vggt} & \second{0.0257} & \third{0.0370} & \second{0.0138} & \second{0.9995} & \best{0.9999} \\
\textbf{\objmodel{}} & \best{0.0149} & \best{0.0243} & \best{0.0079} & \best{0.9996} & \best{0.9999} \\
\bottomrule
\end{tabular}}
\end{minipage}\hfill
\begin{minipage}[t]{0.49\linewidth}
\centering
\textbf{Image-to-3D (full geometry)}\\[-0.4ex]
\resizebox{1.00\linewidth}{!}{%
\begin{tabular}{lccccc}
\toprule
Method & L1$\downarrow$ & L2$\downarrow$ & F@0.01$\uparrow$ & F@0.05$\uparrow$ & Output \\
\midrule
TRELLIS.2~\citep{trellis2} & 0.0566 & 0.00717 & 0.204 & 0.598 & Mesh \\
SAM 3D~\citep{sam3d} & \third{0.0475} & \second{0.00501} & 0.203 & 0.675 & 3DGS \\
LaS-Comp~\citep{lascomp} & 0.0477 & 0.00739 & 0.225 & \third{0.677} & Mesh \\
ReconViaGen~\citep{reconviagen} & 0.0478 & \third{0.00528} & \third{0.228} & \third{0.677} & Mesh \\
\objmodel{}* & \second{0.0326} & 0.00530 & \second{0.321} & \second{0.808} & Mesh \\
\textbf{\objmodel{}} & \best{0.0213} & \best{0.00194} & \best{0.549} & \best{0.898} & PC \\
\bottomrule
\end{tabular}}
\end{minipage}
\end{table}

We verify two hypotheses: our pixel-aligned multilayer paradigm yields (i)~more faithful 3D geometry across objects, scenes, and dynamic clips, and (ii)~geometry that benefits downstream pipelines depending on pose, correspondence, or disoccluded structure. A long-form supplementary video provides additional qualitative results across all regimes.

\begin{table}[t]
\centering
\setlength{\tabcolsep}{2.6pt}
\caption{\textbf{Scene geometry.} Visible-surface depth and point-cloud geometry after the same per-sample scale--shift alignment for all methods. DA3 denotes Depth Anything 3; Pi3X is the upgraded $\pi^3$ model; \scenemodel{}* denotes a scene model trained only on 3D-FRONT. CD-L1, CD-L2, and F-score are computed after unprojection; All-L metrics are only defined for multilayer methods. \small \textcolor{rankone}{\rule{0.9em}{0.9em}} Best \quad
\textcolor{ranktwo}{\rule{0.9em}{0.9em}} Second best \quad
\textcolor{rankthree}{\rule{0.9em}{0.9em}} Third best.}
\vspace{-2pt}
\label{tab:scene_geometry}
\resizebox{1.00\linewidth}{!}{%
\begin{tabular}{lccccccccc}
\toprule
Method & MAE$\downarrow$ & RMSE$\downarrow$ & AbsRel$\downarrow$ & L0 CD-L1$\downarrow$ & L0 CD-L2$\downarrow$ & L0 F@0.05$\uparrow$ & All-L CD-L1$\downarrow$ & All-L CD-L2$\downarrow$ & All-L F@0.05$\uparrow$ \\
\midrule
\multicolumn{10}{c}{\emph{3D-FRONT (held-out, 50 samples)}} \\
\midrule
VGGT~\citep{vggt} & 0.0393 & 0.0568 & 0.0441 & 0.0345 & 0.002889 & 0.7839 & -- & -- & -- \\
DA3~\citep{depthanything3} & 0.0369 & 0.0521 & 0.0403 & 0.0330 & 0.002893 & 0.7992 & -- & -- & -- \\
LaRI-scene~\citep{lari} & 0.0319 & 0.0483 & 0.0359 & 0.0268 & 0.001994 & 0.8576 & \third{0.0575} & \third{0.009990} & \third{0.6671} \\
MoGe-2~\citep{moge2} & 0.0248 & \third{0.0373} & 0.0269 & 0.0213 & 0.000973 & 0.9131 & -- & -- & -- \\
Pi3X~\citep{pi3} & \third{0.0234} & 0.0375 & \third{0.0260} & \third{0.0192} & \third{0.000871} & \third{0.9206} & -- & -- & -- \\
\scenemodel{}* & \second{0.0106} & \second{0.0220} & \second{0.0118} & \second{0.0097} & \second{0.000230} & \second{0.9845} & \second{0.0224} & \second{0.001350} & \second{0.8890} \\
\textbf{\scenemodel{}} & \best{0.0102} & \best{0.0215} & \best{0.0114} & \best{0.0093} & \best{0.000204} & \best{0.9867} & \best{0.0216} & \best{0.001255} & \best{0.8951} \\
\midrule
\multicolumn{10}{c}{\emph{Internal test set}} \\
\midrule
VGGT~\citep{vggt} & 0.0546 & 0.0835 & 0.0779 & 0.0452 & 0.017798 & 0.7502 & -- & -- & -- \\
LaRI-scene~\citep{lari} & 0.0542 & 0.0837 & 0.0769 & 0.0429 & 0.022125 & 0.7591 & \second{0.0872} & \second{0.052319} & \second{0.5565} \\
DA3~\citep{depthanything3} & 0.0462 & 0.0727 & 0.0629 & 0.0387 & 0.005977 & 0.7772 & -- & -- & -- \\
MoGe-2~\citep{moge2} & \third{0.0395} & \third{0.0637} & \third{0.0561} & \third{0.0315} & \third{0.003969} & \third{0.8375} & -- & -- & -- \\
Pi3X~\citep{pi3} & \second{0.0359} & \second{0.0622} & \second{0.0514} & \second{0.0304} & \second{0.003784} & \second{0.8479} & -- & -- & -- \\
\textbf{\scenemodel{}} & \best{0.0328} & \best{0.0586} & \best{0.0470} & \best{0.0278} & \best{0.002850} & \best{0.8720} & \best{0.0750} & \best{0.045000} & \best{0.6500} \\
\bottomrule
\end{tabular}
}
\end{table}

\begin{figure}[t]
  \centering
  \includegraphics[width=\linewidth]{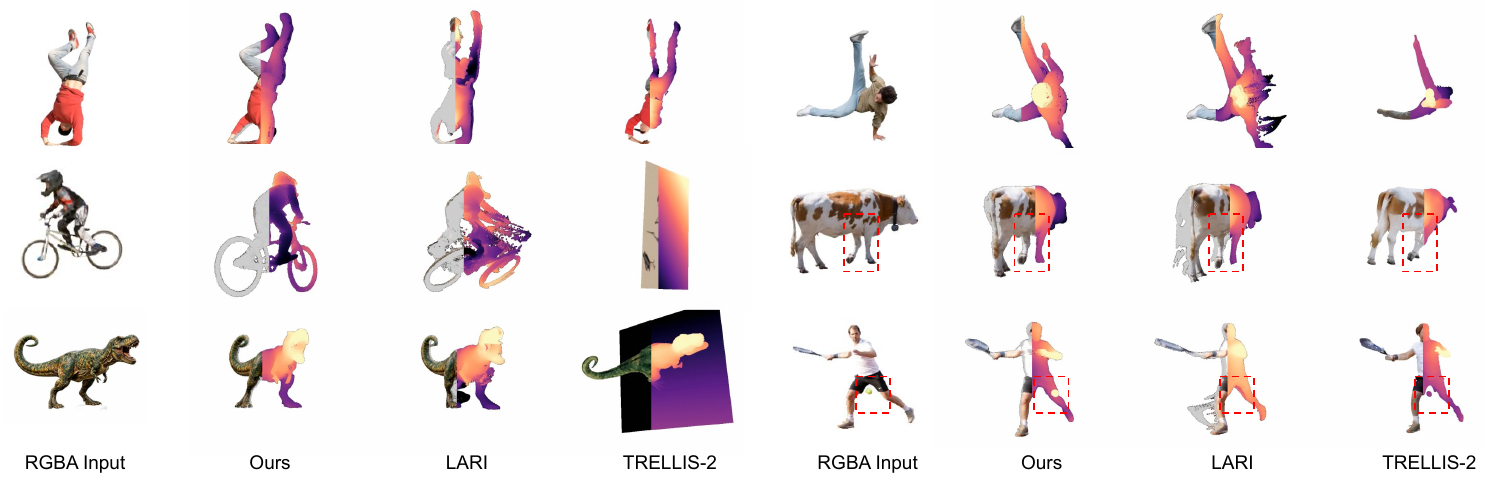}
  \includegraphics[width=\linewidth]{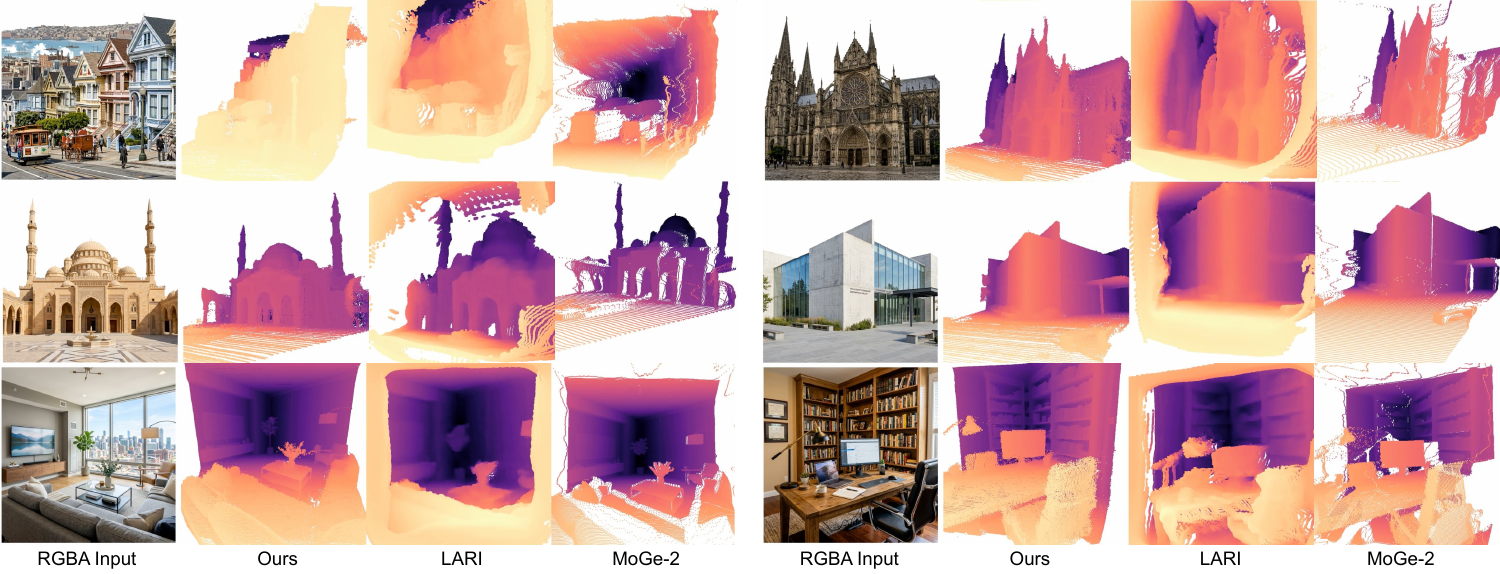}
  \vspace{-5pt}
  \caption{\textbf{Qualitative comparison on out-of-distribution inputs.} All inputs are deliberately drawn from outside our training distributions to probe generalization. Top: object examples comparing \objmodel{} against LaRI-O and TRELLIS.2; inputs are real-world DAVIS~\citep{davis} video frames and generated images, neither of which appear in our object training corpus. Bottom: scene examples comparing \scenemodel{} against LaRI-S and MoGe-2 on generated room images that lie outside the 3D-FRONT scene training distribution. \method{} preserves pixel alignment while generating complete geometry, improving real-image generalization and avoiding several common failure modes of existing SOTAs.}
  \label{fig:qualitative_comparison}
\end{figure}
\subsection{Experimental Setup}
\label{sec:exp_data}

\textbf{Data.}
The multilayer 3D-asset training mixture has three regimes: \textasciitilde300K objects (\textasciitilde17M rendered views) from public collections including Objaverse-XL~\citep{objaversexl}, Objaverse~\citep{objaverse}, 3D-FUTURE~\citep{3dfuture}, Toys4k~\citep{toys4k}, GSO~\citep{gso}, and TrueBones~\citep{truebones}; scene frames from the public 3D-FRONT corpus~\citep{3dfront} plus an internal scene corpus; and \textasciitilde16.8K animated assets sampled for \dynmodel{} (Objaverse-XL animated subset plus Truebones~\citep{truebones} rigged characters). In addition, the mix-training paradigm of Sec.~\ref{sec:mixtrain} lets \scenemodel{} also consume a 12-dataset single-layer \emph{RGBD-style corpus} (real photographs from ScanNet~v2~\citep{scannet}, MegaDepth~\citep{megadepth}, BlendedMVS~\citep{blendedmvs}, ArkitScenes~\citep{arkitscenes}, Argoverse2~\citep{argoverse2}, and Waymo~Open~\citep{waymo}, plus large synthetic single-view sets including Hypersim~\citep{hypersim}, Taskonomy~\citep{taskonomy}, and three smaller datasets), supervised on $L_0$ only via Eq.~\ref{eq:singlelayer_mask}. Every result reported in this section (Tables~\ref{tab:object_results}--\ref{tab:timestep_sampling}) uses the 3D-asset-only \method{} checkpoints with no RGBD data in the mix; only the real-scene depth comparison in App.~\ref{app:real_scene_depth} (Table~\ref{tab:real_scene_depth}) reports the mix-trained \scenemodel{}, where we list it as a separate row alongside the 3D-asset-only baseline so the contribution of the RGBD mix is visible as an internal ablation. Source datasets, rendering details, and the full RGBD-style corpus specification are in App.~\ref{app:data} and App.~\ref{app:data_rgbd}.

\textbf{Training details.}
\methodnet{} is a $1.7$B-parameter ($1.4$B trainable) DiT trained at $504{\times}504$ resolution with $L{=}6$ layers, optimized with AdamW on $64$ H100 GPUs (global batch size $512$); \dynmodel{} is fine-tuned from the \objmodel{} checkpoint. Inference uses $20$ denoising steps. See App.~\ref{app:training} for details.

\textbf{Evaluation.}
We evaluate primarily on reproducible held-out public-data benchmarks: $100$ held-out object assets, the held-out 3D-FRONT split for scenes, and Obj.-Val / Truebone / ActionBench for dynamic clips. We additionally report a $200$-sample internal scene test set only as a generalization probe beyond furnished rooms. Baselines cover dedicated depth predictors, layered predictors, image-to-3D generators, dynamic-geometry methods, and TRELLIS hybrids; for generative methods we draw $K{=}8$ random seeds per input and report the best geometry result. Full split definitions and baseline lists are in App.~\ref{app:data}.

\textbf{Metrics.}
We emphasize geometry faithfulness over aesthetic plausibility: standard depth errors on visible surfaces and Chamfer/F-score on complete geometry, with visible and occluded breakdowns where applicable. For downstream pipelines, the TRELLIS hybrid is measured through the same geometry metrics, while scene editing and view synthesis are presented as qualitative demonstrations in the paper and supplemental video. Detailed metric definitions are in App.~\ref{app:data}.

\subsection{Faithful Geometry Generation}
\label{sec:exp_object}

\textbf{Objects.}
Table~\ref{tab:object_results} evaluates the object model and Fig.~\ref{fig:qualitative_comparison} (top) shows qualitative comparisons on out-of-distribution inputs (DAVIS~\citep{davis} video frames and generated images). 
Although \objmodel{} predicts six layers, its layer 0 achieves the best visible-surface accuracy among all depth and pointmap baselines on MAE, RMSE, and AbsRel: generation of occluded geometry does not cost visible-surface faithfulness. \objmodel{} also achieves the best complete-object geometry in this benchmark, improving over textured-mesh and 3D-Gaussian generators, and further improves TRELLIS.2 when used as its Stage-1 prior.
The qualitative gains stem from two complementary factors over prior work: a pixel-aligned diffusion (vs.\ LaRI's regression head) better models the multi-modal uncertainty of occluded surfaces and avoids deep-layer mask collapse, and inheriting 2D image priors from the frozen MoGe encoder generalizes better to real images than canonical-frame asset-trained generators that can flip pose or hallucinate spurious planes (App.~\ref{app:why_objects}).

\begin{figure*}[t]
  \centering
  \includegraphics[width=1\textwidth]{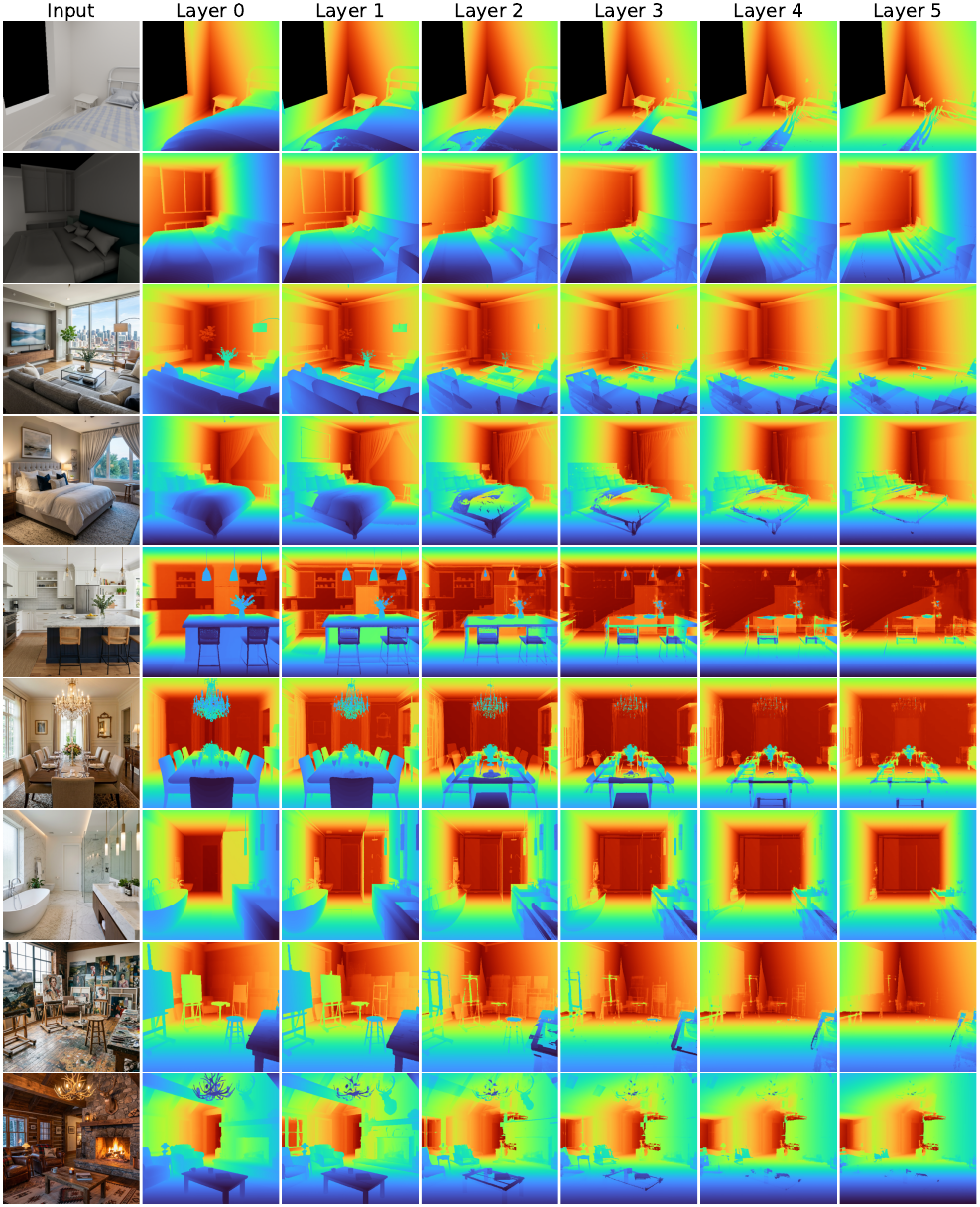}
  \vspace{-2pt}
  \caption{\textbf{Multilayer depth stack produced by \scenemodel{}.} Each row shows one input (left) followed by the six predicted depth layers in turbo colormap. Top two rows are held-out 3D-FRONT frames; the remaining seven are out-of-distribution generated indoor rooms. As $\ell$ increases, occluded geometry behind near surfaces is filled in (e.g.\ floor and walls behind furniture, room interiors behind doorways and chandeliers) while Layer~0 stays pixel-aligned with the input. Predictions use the \scenemodel{} checkpoint with $20$ denoising steps.}
  \label{fig:multilayer_depth_gallery}
\end{figure*}

\textbf{Scenes.}
Table~\ref{tab:scene_geometry} reports scene geometry performance on the reproducible held-out 3D-FRONT benchmark and a 200-sample internal scene test set separately as a generalization probe, with qualitative examples on out-of-distribution generated images in Fig.~\ref{fig:qualitative_comparison}.
After the same SSI alignment used for all depth baselines, \scenemodel{} achieves the best visible-surface depth and L0 point-cloud geometry among the compared methods, and improves substantially over LaRI-scene on all-layer geometry, where single-layer baselines do not produce occluded layers.
The 3D-FRONT-only \scenemodel{}* nearly matches \scenemodel{} on held-out 3D-FRONT, isolating the effect of the broader internal scene corpus. Despite mainly training on virtual scenes and without MoGe-2-scale real-scene data, \scenemodel{} is more faithful on out-of-distribution generated images, e.g., preserving planar facades and walls that MoGe-2 bends (App.~\ref{app:why_scenes}). See App.~\ref{app:real_scene_depth} for depth estimation results on NYU Depth V2~\citep{nyuv2} and ETH3D~\citep{eth3d}. Figure~\ref{fig:multilayer_depth_gallery} shows the full $L{=}6$ layer stack produced by \scenemodel{} on held-out 3D-FRONT frames and out-of-distribution generated rooms; deeper layers progressively populate occluded geometry (e.g.\ floor and back-wall regions behind furniture) while the visible Layer~0 remains faithful to the input.

\textbf{Dynamic clips.}
Table~\ref{tab:dynamic_geometry} evaluates \dynmodel{} against GVFDiffusion (GVFD), SS4D, and ActionMesh (AM) on three complementary splits. Obj.-Val uses held-out Objaverse-XL animated objects, Truebone uses rigged assets with explicit articulation, and ActionBench focuses on action-driven motions. \dynmodel{} wins on Obj.-Val and Truebone and achieves the best mean CD-L2. AM is strongest on ActionBench, where ground truth is tracked animated surfaces matching its native output format.


\begin{table}[t]
\begin{minipage}[t]{0.52\linewidth}
\vspace{0pt}
\centering
\setlength{\tabcolsep}{2.6pt}
\caption{\textbf{Dynamic geometry.} Global CD-L2 ($\downarrow$).}
\vspace{-5pt}
\label{tab:dynamic_geometry}
\resizebox{0.90\linewidth}{!}{%
\begin{tabular}{lcccc}
\toprule
Benchmark & GVFD~\citep{gvf_diffusion} & SS4D~\citep{ss4d} & AM~\citep{actionmesh} & \dynmodel{} \\
\midrule
ActionBench & \third{0.0879} & 0.0882 & \best{0.0243} & \second{0.0291} \\
Truebone & 0.0166 & \third{0.0145} & \second{0.0120} & \best{0.0063} \\
Obj.-Val & \third{0.0248} & 0.0249 & \second{0.0142} & \best{0.0034} \\
Mean & 0.0385 & \third{0.0381} & \second{0.0162} & \best{0.0105} \\
\bottomrule
\end{tabular}}
\end{minipage}\hfill
\begin{minipage}[t]{0.48\linewidth}
\vspace{0pt}
\centering
\setlength{\tabcolsep}{4pt}
\caption{\textbf{Timestep sampling ablation.} CD-L2.}
\vspace{-5pt}
\label{tab:timestep_sampling}
\resizebox{0.85\linewidth}{!}{%
\begin{tabular}{lccc}
\toprule
Schedule & L0 & L1--L5 & All \\
\midrule
Plat. logit-normal & \second{0.021} & \third{0.033} & \third{0.031} \\
Logit-normal & \third{0.023} & \second{0.028} & \second{0.027} \\
\textbf{Mixture} & \best{0.020} & \best{0.025} & \best{0.024} \\
\bottomrule
\end{tabular}}
\end{minipage}
\end{table}

\subsection{Discussion}
\label{sec:exp_abl}

The experiments above measure final geometry quality. Here we discuss the design choices that make the representation trainable and robust; an extended discussion of remaining design choices, generalization, and limitations is in App.~\ref{app:discussion_extra}.

\textbf{Depth filling rather than mask prediction.}
\label{abl:mask}
A direct multilayer formulation couples two tasks: regressing XYZ and classifying whether each layer exists. In early joint depth+mask runs, EMA cosine similarity between the two gradients stays near zero or negative (e.g., $-0.19$ at $500$ iters), strongest in the early decoder blocks. Mask supervision is also severely imbalanced---on object renderings the mean valid area drops from $8.14\%$ at L0 to $0.60\%$ at L5 (App.~\ref{app:layer_stats})---making the trivial ``invalid'' prediction attractive and causing mask heads to suppress deeper surfaces. Forward filling converts this into a dense XYZ regression: a diagnostic shows originally valid deeper-layer pixels are only slightly higher-error than inherited ones, both within the same regime, confirming filling as an effective dense supervision strategy.

\textbf{Timestep sampling.}
\method{} sits in a natural progression of pixel-aligned geometry models: MoGe regresses a single visible layer, PPD~\citep{ppd} formulates single-layer geometry as diffusion with a uniform timestep schedule, and \method{} extends this direction to multilayer flow matching. The new ingredient is a layer-aware schedule mixture: the visible layer is closer to faithful visible-surface prediction, while occluded layers are closer to conditional generation, so a single diffusion-time distribution is suboptimal. Table~\ref{tab:timestep_sampling} compares standard logit-normal, plateaued logit-normal, and the final mixture schedule. Plateaued sampling helps L0 but hurts deeper generative layers; the mixture gives the best overall stack quality by balancing the two regimes.


\section{Downstream Pipeline Demonstrations}
\label{sec:applications}

We demonstrate three downstream uses of \method{} to show how the geometric advantages measured in Sec.~\ref{sec:experiments} transfer to real pipelines. Figure~\ref{fig:pc_mesh_video} summarizes the central pattern: because \method{} outputs complete geometry in the input camera frame while preserving intrinsics and pixel-level correspondence, this representation enables object insertion, textured mesh generation, and geometry-guided video synthesis. For TRELLIS hybridization, Table~\ref{tab:object_results} shows that our pixel-aligned point stack is a stronger Stage-1 prior than pure TRELLIS.2 or VGGT-guided alternatives. The resulting mesh remains aligned with the input image and recovered camera intrinsics, unlike canonical-frame meshes that must be aligned after generation. For view synthesis, complete multilayer geometry serves as memory for a video model, giving it explicit support for dis-occluded regions under large viewpoint changes. The long-form demo video provides many additional visual examples across these downstream uses, including object insertion, multi-object scene edits, and view-synthesis videos. All three pipelines below reuse the same \method{} prediction (camera-space multilayer point stack with recoverable intrinsics) and add only lightweight, training-free composition logic.

\subsection{Text-Driven 3D Scene Editing}
\label{sec:app_insert}

Given a photograph and a natural-language edit, a 2D editor can produce a plausible edited image, but not a 3D-consistent scene. \method{} closes this gap. As shown by the object-insertion example in the final panel of Fig.~\ref{fig:teaser}, we first rely on the 2D editor's ability to add, remove, or replace image content, then lift the edited result into 3D while preserving image-grid correspondence. We predict the original scene with \scenemodel{}, predict the generated object or edited region with \objmodel{}, and insert it into the 3D scene in closed form because both outputs already live in the photograph's camera frame and share pixel/point correspondences. The resulting agent supports object insertion, removal, replacement, and compositional edits from a single image, without per-edit optimization or a rendering loop. The long-form demo video provides many additional examples. This is precisely where pixel alignment matters: a canonical-frame object generator may produce a plausible asset, but it does not know where that asset lies in the input camera.

\subsection{Geometry-Guided Novel-View Video Synthesis}
\label{sec:app_video}

Figure~\ref{fig:pc_mesh_video} illustrates this view-synthesis path. Image-to-video world models often use a depth map or rendered geometry track as conditioning~\citep{wonderland, trajectorycrafter, freeorbit4d, jiang2025vaceallinonevideocreation}. A single visible layer, however, is incomplete by construction: once the camera moves, newly exposed regions must be invented from scratch. \method{} provides a complete geometry memory before video generation begins. Crucially, because \method{}'s multilayer geometry is pixel-aligned with the input image by construction, it integrates with these video models: we simply rasterize the predicted point cloud along a target trajectory to generate dense, multi-view depth guidance. For object orbits, the back side is already present, allowing the frozen video model to edit and fill a geometry-consistent signal rather than hallucinating disoccluded structure without support.

\subsection{Pose-Aligned TRELLIS Hybrid}
\label{sec:app_trellis}

Figure~\ref{fig:pc_mesh_video} also shows the pose-aligned mesh path. TRELLIS-style pipelines produce visually polished textured meshes, but their sparse structure is generated in a canonical frame and has no explicit input camera pose~\citep{trellis}. The mesh may look plausible while failing to reproject to the image that conditioned it. Since \method{} already predicts a camera-space, pixel-aligned point stack with recoverable intrinsics, we voxelize this stack and use it as the sparse structure for the later TRELLIS stages, without retraining TRELLIS. The hybrid keeps TRELLIS' texture and mesh decoder while replacing its weakest link: the pose-agnostic Stage-1 structure. We evaluate against pure TRELLIS/TRELLIS-2 and VGGT-guided hybrids such as ReconViaGen~\citep{reconviagen} and LaS-Comp~\citep{lascomp}.

\section{Conclusion}

We presented \methodlong{}, a pixel-aligned multilayer geometry representation for faithful generation, where visible-surface fidelity and occluded-geometry completion become successive layers of one camera-space tensor. \methodnet{} learns this representation with a frozen 2D foundation encoder, flow matching, and an XYZ-only depth-filling objective, and applies across objects, scenes, and dynamic clips. The resulting geometry is not only accurate in benchmarks; it also provides a stronger 2D-to-3D interface for scene editing, view synthesis, and pose-aligned textured mesh generation. We hope \method{} becomes a common substrate for 3D-aware perception, generation, editing, and video.


\clearpage
\bibliographystyle{plainnat}
\bibliography{references}

\appendix
\clearpage
\section*{Technical Appendices and Supplemental Material}
\addcontentsline{toc}{section}{Technical Appendices and Supplemental Material}

\noindent
This supplementary document provides details deferred from the main paper:
\begin{itemize}[leftmargin=*,itemsep=2pt,topsep=4pt]
    \item App.~\ref{app:method} -- method details (scale normalization, invalid-pixel noise fill, monotonicity penalty).
    \item App.~\ref{app:training} -- training schedule, optimizer, model hyperparameters, decoder details, and the layer-aware diffusion-time schedule.
    \item App.~\ref{app:data} -- rendering pipeline, source datasets, evaluation splits, baselines, and metric definitions.
    \item App.~\ref{app:layer_stats} -- per-layer validity statistics on the object corpus.
    \item App.~\ref{app:why_better} -- qualitative analysis of why \objmodel{} and \scenemodel{} outperform their respective baselines.
    \item App.~\ref{app:discussion_extra} -- extended discussion on layer count, prediction target, architecture details, generalization, limitations, and future work.
\end{itemize}

\section{Method Details}
\label{app:method}

This appendix expands Sec.~\ref{sec:formulation} with details deferred from the main paper.

\textbf{Scale normalization.}
\label{app:scale_norm}
Raw camera-space coordinates span very different ranges for objects and scenes, so we normalize XYZ with a reversible map chosen per regime. For object and dynamic-object data, whose scale is controlled by rendering, we use a global per-channel z-score
\begin{equation}
\tilde{\mathbf{x}} \;=\; (\mathbf{x} - \boldsymbol{\mu})\,/\,\boldsymbol{\sigma},
\label{eq:zscore}
\end{equation}
where $\boldsymbol{\mu},\boldsymbol{\sigma}\in\R^3$ are channel-wise statistics computed once on the training corpus. For scenes, where depth can vary by orders of magnitude within a trajectory, we use a per-sample log-median map. Let $m$ be the median valid depth over the multilayer stack:
\begin{equation}
\tilde z \;=\; \ln(z/m), \qquad
\tilde x \;=\; \mathrm{sign}(x)\,\ln(1 + |x|/m), \qquad
\tilde y \;=\; \mathrm{sign}(y)\,\ln(1 + |y|/m).
\label{eq:medianlog}
\end{equation}
Both maps are reversible, so predictions can be returned to camera-space metric coordinates at inference. The same architecture predicts the normalized tensor in both regimes; only the input/output coordinate transform changes.

\textbf{Invalid input pixels and noise fill.}
\label{app:invalid_pixels}
The input image contains invalid pixels outside the layer-$0$ silhouette support (e.g., the background of a segmented object, or sky pixels in a scene). Let $A(\mathbf{u})\in\{0,1\}$ denote the layer-$0$ alpha validity, broadcast across all $L$ layers and XYZ channels. These invalid pixels are excluded from the endpoint loss in Eq.~\ref{eq:fm}. During both training and inference, the noisy geometry feature at invalid pixels is replaced by fresh Gaussian noise before patchification:
\begin{equation}
\mathbf{x}_t^{\mathrm{net}}(\ell,\mathbf{u}) \;=\; A(\mathbf{u})\cdot \mathbf{x}_t(\ell,\mathbf{u}) \;+\; (1 - A(\mathbf{u}))\cdot \boldsymbol{\epsilon}(\ell,\mathbf{u}),\qquad \boldsymbol{\epsilon}\sim\mathcal{N}(\mathbf{0},\mathbf{I}).
\label{eq:noisefill}
\end{equation}
This invalid-pixel corruption tells the network to ignore geometry tokens at invalid pixels, while the dense forward fill (Eq.~\ref{eq:forwardfill}) defines what the network should predict along valid input rays. Together they create a dense XYZ prediction problem with the layer-$0$ alpha as the only visibility input and no auxiliary per-layer mask head: valid input rays always have a target for every layer, and invalid pixels carry only uninformative noise and do not contribute to the endpoint loss.

\textbf{Monotonicity penalty.}
\label{app:mono}
We add a soft adjacent-layer monotonicity penalty in the normalized coordinate space. Because the object/dynamic z-score and scene log-median depth transform are monotone in depth, this penalty preserves the same front-to-back ordering as camera-space depth and is active only when adjacent layers violate that order:
\begin{equation}
\mathcal{L}_{\mathrm{mono}} \;=\; \frac{1}{L-1}\sum_{\ell=0}^{L-2}\E_\mathbf{u}\!\bigl[\,\relu\!\bigl(z_\ell(\mathbf{u}) - z_{\ell+1}(\mathbf{u})\bigr)^2\bigr],
\label{eq:mono}
\end{equation}
where $z_\ell(\mathbf{u})$ is the normalized depth coordinate of the predicted layer-$\ell$ point at pixel $\mathbf{u}$. The full training loss is $\mathcal{L}=\mathcal{L}_{\mathrm{FM}}+\lambda_{\mathrm{mono}}\mathcal{L}_{\mathrm{mono}}$ with $\lambda_{\mathrm{mono}}{=}0.1$. The penalty is one-sided (zero when ordering is correct) and acts as a gentle structural prior; we found it stabilizes early training without measurably affecting final geometry quality.

\section{Training Details}
\label{app:training}

This appendix expands the \emph{Training details} paragraph in Sec.~\ref{sec:exp_data} with the full hyperparameter list and recipe.

\textbf{Backbone and resolution.}
Object and dynamic-object models are trained at $504\!\times\!504$ resolution with patch size $14$ and $L{=}6$ layers. The frozen MoGe ViT-L~\citep{moge} encoder supplies image tokens: we aggregate features from its last four blocks and project them to the decoder dimension, and only this projection is updated; the rest of the encoder is never trained. The object backbone uses $48$ pre-norm DiT blocks~\citep{sd3} at width $D{=}1536$ with $24$ attention heads. At every $(\ell,\mathbf{u})$ token the layer-$\ell$ noisy geometry is concatenated channel-wise with the (repeated) image feature at pixel $\mathbf{u}$ and projected to $D$, producing a $(B,L,P,D)$ token grid where $P$ is the number of $14\!\times\!14$ patches. The decoder cycles through layer-wise, ray-wise, and global attention shapes (Sec.~\ref{sec:architecture}) followed by a linear projection from each decoder token to the XYZ $x_0$ of its patch and unpatchification to the full multilayer image grid. The model contains about $1.7$B parameters in total, of which about $1.4$B are trainable after freezing MoGe.

\textbf{Temporal attention block.}
For \dynmodel{}, a clip of $T{=}16$ frames is flattened along $(T\!\cdot\!L)$, and one temporal-attention block is inserted after each global-attention block. Each temporal block reshapes tokens to $(B\!\cdot\!L\!\cdot\!P,\,T,\,D)$, applies a single self-attention layer along $T$ with 1D RoPE on the time axis, and reuses the host decoder's time AdaLN. The block is initialized with LayerScale~\citep{layerscale} $\gamma{=}10^{-5}$ on its residual, so warm-starting \dynmodel{} from a single-frame \objmodel{} checkpoint reproduces the static behaviour bit-for-bit on $T{=}1$ inputs and only gradually picks up temporal coupling during fine-tuning.

\textbf{Training schedule.}
We train the object model in two stages: a $196\!\times\!196$ low-resolution stage for $100$K iterations for faster representation learning, followed by $100$K iterations at $504\!\times\!504$. The dynamic model is initialized from the static object checkpoint, augmented with temporal attention blocks, and fine-tuned on clips for another $50$K iterations.

\textbf{Optimizer.}
AdamW with learning rate $10^{-4}$, minimum learning rate $10^{-5}$ under cosine decay, $2000$ warmup iterations, weight decay $0.01$, and betas $(0.9,0.999)$. The main object run uses $64$ H100 GPUs with per-GPU batch size $2$ and gradient accumulation $4$, giving an effective batch size of $512$. Gradients are clipped to norm $1.0$. EMA weights use decay $0.9995$ after warmup. Rare loss spikes above $4{\times}$ the running EMA are skipped to avoid corrupting Adam statistics.

\textbf{Inference.}
We solve the flow ODE with $20$ denoising steps. For object and dynamic models the predicted normalized tensor is mapped back to camera-space XYZ via the inverse z-score; for scene models, via the inverse log-median map (App.~\ref{app:scale_norm}).

\textbf{Robust silhouettes.}
Real input masks are imperfect: they may shift boundaries, miss thin parts, or leave small holes inside the visible support. During training we jitter the alpha boundary and treat newly exposed or missing-support pixels as pseudo-labeled regions. These pixels inherit the filled geometry target and are supervised at reduced weight, rather than being treated as fully reliable surfaces. This improves boundary and mask-error robustness without introducing a separate silhouette loss, and explains why the same checkpoint remains stable in the TRELLIS hybrid and scene-editing pipelines, which depend on externally produced masks.

\textbf{Self-consistent intrinsics.}
For images with unknown $K$, we recover a self-consistent pinhole intrinsics matrix directly from the predicted layer-$0$ pointmap by least-squares fitting the projection equations, following MoGe~\citep{moge}. Concretely, with predicted camera-space points $\mathbf{x}_0(\mathbf{u})=(X,Y,Z)$ at image pixels $\mathbf{u}=(u,v)$, we solve for $(f_x,f_y,c_x,c_y)$ by minimizing
\[
\sum_\mathbf{u}\Bigl\|\bigl(u,v\bigr)-\bigl(f_x X/Z+c_x,\; f_y Y/Z+c_y\bigr)\Bigr\|^2
\]
over valid layer-$0$ pixels. The recovered intrinsics are self-consistent with the predicted pointmap and input pixel grid, which is essential for camera-aware downstream uses such as object insertion and novel-view video synthesis.

\begin{figure}[h]
  \centering
  \includegraphics[width=0.7\linewidth]{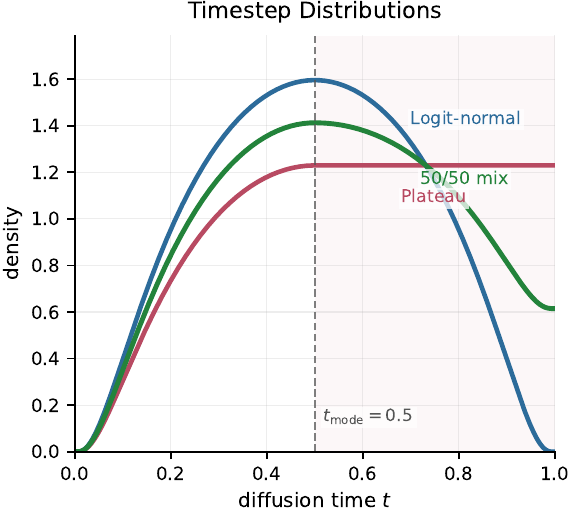}
  \caption{\textbf{Training timestep distributions.} Analytic densities of the standard logit-normal schedule, the plateaued logit-normal variant, and their equal-weight 50/50 mixture used in phase~(ii) of the layer-aware curriculum.}
  \label{fig:timestep_schedules}
\end{figure}

\textbf{Layer-aware diffusion-time schedule.}
\label{app:noise_curriculum}
The training noise curriculum summarized in Sec.~\ref{sec:impl} uses two complementary timestep distributions over $t\in[0,1]$. The standard \emph{logit-normal} schedule~\citep{sd3} concentrates samples near $t{=}0.5$ and is well suited to deeper layers, where the prediction problem is closer to image-conditioned 3D generation. The \emph{plateaued logit-normal} variant inspired by representation-comparison studies~\citep{bfl_representation} broadens the central plateau and allocates more probability to small $t$, which favors layer~$0$ where the prediction is closer to faithful visible-surface reconstruction. Training proceeds in two phases: (i)~early in optimization, diffusion times are sampled \emph{independently} per layer (visible layer from the plateaued logit-normal, deeper layers from the standard logit-normal), so each layer is exposed primarily to the noise regime that matches its uncertainty profile; (ii)~once all layers have stabilized, the full stack switches to a \emph{shared} timestep drawn from an equal-weight mixture of the two distributions, which keeps coupling across layers consistent at inference. Figure~\ref{fig:timestep_schedules} plots the analytic densities of the two base schedules and their 50/50 mixture used in phase~(ii).

\section{Data Pipeline Details}
\label{app:data}

This appendix expands Sec.~\ref{sec:rendering} with the full data-pipeline specification.

\textbf{Rendered inputs and source datasets.}
%
We render RGBA images with randomized lighting, viewpoints, and intrinsics. \emph{Object data} covers roughly $300K$ unique assets and $17M$ rendered views drawn from Objaverse-XL~\citep{objaversexl}, Objaverse~\citep{objaverse}, 3D-FUTURE~\citep{3dfuture}, Toys4k~\citep{toys4k}, GSO~\citep{gso}, and rigged characters from Truebones~Motions Animation Studios~\citep{truebones} rendered as static views. \emph{Scene data} includes 3D-FRONT~\citep{3dfront} indoor rooms and an additional internal curated scene corpus with varied intrinsics. \emph{Dynamic data} covers roughly $16.8K$ animated assets sampled as short clips, used to train and evaluate \dynmodel{}, combining held-in subsets of Objaverse-XL animated assets with rigged characters from Truebones~\citep{truebones}; held-out splits of the same sources form the Obj.-Val and Truebone evaluation benchmarks.

\textbf{RGBD-style mix-training corpus.}
\label{app:data_rgbd}
The mix-training paradigm of Sec.~\ref{sec:mixtrain} additionally exposes \scenemodel{} to a $12$-dataset RGBD-style corpus that provides only the visible-surface layer $L_0$. Following Eq.~\ref{eq:singlelayer_mask}, every frame in this corpus carries $b_{\mathrm{single}}{=}1$, so the supervision contributes to $L_0$ only and the network's predictions for $L_1,\dots,L_5$ remain shaped exclusively by the multilayer 3D-asset corpus. The constituent datasets group naturally into two families that match what dedicated monocular-depth predictors consume:

\begin{itemize}[leftmargin=*,itemsep=0pt,topsep=2pt]
\item \emph{Real RGBD photographs.}
ScanNet~v2~\citep{scannet} ($\sim$2.5M structured-light RGB-D frames from $1{,}513$ indoor scans);
MegaDepth~\citep{megadepth} (Internet photo collections with MVS-derived depth on $\sim$150K landmark images);
BlendedMVS~\citep{blendedmvs} ($\sim$17K frames with photometrically blended MVS depth across diverse object- and scene-scale captures);
ArkitScenes~\citep{arkitscenes} ($\sim$5K rooms with Apple ARKit ToF depth and laser-scanner ground truth);
Argoverse2~\citep{argoverse2} (LiDAR-derived sparse depth on $1{,}000$ outdoor driving scenes);
and Waymo~Open~\citep{waymo} (LiDAR-derived sparse depth on $\sim$1{,}150 outdoor driving segments).
\item \emph{Synthetic single-view RGBD.}
Hypersim~\citep{hypersim} ($\sim$77K photorealistic indoor renders with dense depth);
Taskonomy~\citep{taskonomy} ($\sim$4.6M indoor frames with dense depth from $\sim$500 buildings);
and three smaller synthetic single-view datasets (Aria Synthetic Environments, ParallelDomain, TartanAir~v2) used as additional indoor and outdoor diversity.
\end{itemize}

For each mini-batch we draw from the 3D-asset corpus with probability $p_{\mathrm{dojo}}{=}0.6$ and from the RGBD-style corpus with $p_{\mathrm{rgbd}}{=}0.4$. Within the RGBD branch, datasets are weighted by $\sqrt{N_{\mathrm{rows}}}$ with a $1.5{\times}$ boost on the six real-photograph sets. Frames are loaded at the same $504{\times}504$ resolution as the rendered corpus and pass through the same online augmentation (random crop, resize, horizontal flip, photometric jitter, no per-frame rotation for sparse-depth driving frames). Depth values are converted to camera-space $XYZ$ with the original intrinsics, normalized by the same per-sample log-median map used for scene data (App.~\ref{app:scale_norm}), and broadcast across $L{=}6$ layers; the $b_{\mathrm{single}}{=}1$ flag then masks all but $L_0$ from the endpoint loss. No new model parameters, head, or regime embedding are added; only the data-side mask gate of Eq.~\ref{eq:singlelayer_mask} differs between the two regimes.

\textbf{Evaluation splits and protocol.}
For object evaluation, we randomly sample $100$ unique assets from the held-out object corpus and evaluate all rendered views for those assets. For scene evaluation, the reproducible benchmark is a held-out 3D-FRONT split; the $200$-sample internal test set is reported separately to probe generalization beyond furnished rooms. For dynamic-object evaluation, Obj.-Val denotes held-out Objaverse-XL animated-object assets, Truebone denotes a held-out split of rigged animated assets, and ActionBench evaluates action-driven articulated motion. Baselines cover dedicated depth predictors, layered predictors (LaRI), image-to-3D generators (TRELLIS.2, SAM 3D, LaS-Comp, ReconViaGen), dynamic-geometry methods (GVFDiffusion, SS4D, ActionMesh), and TRELLIS hybrids. For stochastic diffusion or generative methods, including TRELLIS-style baselines and our hybrids, we draw $K{=}8$ random seeds per input and report the best geometry result.

\textbf{Metrics.}
We emphasize geometry faithfulness over aesthetic plausibility, intentionally separating these measurements from mesh or video scores that often reward visual plausibility rather than agreement with the input geometry. For visible surfaces we report standard depth errors (MAE, RMSE, AbsRel, $\delta{<}1.25^k$). For complete geometry we report Chamfer distance (L1, L2) and F-score to ground-truth meshes or point samples, with visible (L0) and occluded (All-L) breakdowns where possible. For dynamic clips we average per-frame geometry errors. For downstream pipelines we measure the property each pipeline needs: image reprojection / pose alignment for TRELLIS hybrids, geometric consistency for view synthesis, and registration quality for scene edits.

\textbf{Multilayer geometry by depth peeling.}
Ground-truth layers are generated with depth peeling~\citep{mammen1989transparency,everitt2001interactive,nvdiffrast}. We render the scene repeatedly from the same camera: the first pass records the nearest visible surface, and each following pass ignores surfaces already captured at smaller depth so that the next intersection along the ray becomes visible. This produces an ordered stack of depth maps, where layer $\ell$ stores the $\ell$-th front-to-back surface hit for each pixel when such a hit exists. We retain the first $L$ surfaces per ray and unproject every valid depth value with the render intrinsics to obtain camera-space XYZ supervision. Rays with fewer than $L$ intersections are handled by the target filling rule in Sec.~\ref{sec:formulation}, giving the network a dense multilayer tensor while preserving the true set of observed intersections.

\textbf{Augmentation.}
We apply online augmentations to narrow the gap between renderings and photos and expand our dataset. Geometric augmentations include random crop, resize, horizontal flip, in-plane rotation, and small affine perturbations such as image-plane translation, anisotropic scale/aspect-ratio jitter, and mild shear, with the camera-space targets transformed consistently with the image operation. Photometric augmentations include brightness, contrast, saturation, hue, gamma, exposure, and white-balance jitter, together with occasional blur, sharpening, compression artifacts, and sensor noise. For alpha masks, we randomly dilate or erode the silhouette, perturb the boundary, drop thin structures, and fill or remove small holes; the affected pixels are handled with the pseudo-labeling strategy described above. For dynamic clips, all geometric and mask-space augmentations are shared across frames to preserve temporal consistency, while photometric noise can vary mildly over time to mimic real video capture.

\section{Layer Validity Statistics}
\label{app:layer_stats}

Table~\ref{tab:layer_validity} reports the per-layer fraction of valid pixels in our object renderings, supporting the discussion in Sec.~\ref{abl:mask} (Depth filling vs.\ mask prediction) and App.~\ref{app:layers_choice} (How many layers?). Coverage is measured over the full $504{\times}504$ rendered image, including background outside the object silhouette. Mean valid coverage drops from $8.14\%$ at L0 to $0.60\%$ at L5, a more than $13\!\times$ imbalance that explains why a per-layer mask head collapses on deeper layers. Six layers cover the overwhelming majority of valid rays while keeping the multilayer tensor compact.

\begin{table}[h]
\centering
\setlength{\tabcolsep}{5pt}
\caption{\textbf{Valid multilayer pixels in object renderings.} Per-layer pixel-coverage statistics aggregated over the object validation corpus, measured over the full rendered image including background. Deeper intersections are much sparser, which makes per-layer mask prediction severely imbalanced.}
\label{tab:layer_validity}
\begin{tabular}{lcccccc}
\toprule
Layer & Mean (\%) & Median (\%) & P25 (\%) & P75 (\%) & P90 (\%) & Valid views \\
\midrule
L0 & 8.14 & 6.21 & 3.92 & 10.56 & 16.19 & 964{,}508 \\
L1 & 7.90 & 6.01 & 3.80 & 10.25 & 15.82 & 963{,}163 \\
L2 & 2.53 & 1.46 & 0.61 & 3.02 & 5.97 & 910{,}810 \\
L3 & 1.74 & 0.86 & 0.29 & 1.93 & 4.20 & 835{,}007 \\
L4 & 0.72 & 0.21 & 0.05 & 0.68 & 1.78 & 696{,}663 \\
L5 & 0.60 & 0.15 & 0.04 & 0.50 & 1.44 & 516{,}342 \\
\bottomrule
\end{tabular}
\end{table}

\section{Geometry Generation: Qualitative Analysis}
\label{app:why_better}

This appendix expands the qualitative discussion deferred from Sec.~\ref{sec:exp_object} (\emph{Objects} and \emph{Scenes}) into the underlying mechanisms.

\subsection{Why \objmodel{} outperforms regression-based and canonical-frame baselines}
\label{app:why_objects}

Two complementary mechanisms drive the object gains in Table~\ref{tab:object_results} and Fig.~\ref{fig:qualitative_comparison} (top).

\textbf{Diffusion vs.\ regression on occluded surfaces.}
Compared with LaRI, which uses a regression-style layered head, our diffusion formulation is better suited to fully invisible back-side geometry: regression tends to average plausible completions, producing smooth or detail-poor backs, whereas diffusion can model the multi-modal uncertainty of occluded surfaces. LaRI also jointly predicts depth and per-layer masks; because deeper-layer supervision is extremely imbalanced (App.~\ref{app:layer_stats}), its mask prediction often collapses after the first few layers, effectively producing only shallow geometry. Our depth-filling objective avoids this failure mode by making all foreground rays dense without requiring a separate mask head.

\textbf{Pixel-aligned representation vs.\ canonical-frame generation.}
Compared with TRELLIS-style generators trained primarily from 3D assets, \objmodel{} generalizes better to real images because its pixel-aligned representation can inherit 2D image priors through the frozen MoGe encoder while still predicting complete geometry. Canonical-frame generators sometimes produce visually polished meshes whose geometry is unfaithful to the input: a plausible-looking mesh can flip front/back leg relationships, while pixel-aligned methods such as LaRI and \objmodel{} preserve the input pose ordering. TRELLIS.2 also occasionally introduces spurious planar structures. These failures are consistent with Table~\ref{tab:object_results}: canonical generators can look attractive, but they are weaker at faithful generation when evaluated against the input geometry and ground truth.

\textbf{Architecture vs.\ data scale.}
A small-scale control with a reduced-layer \objmodel{} and a LaRI-comparable data budget shows the same qualitative trend, suggesting that the gains are not solely a consequence of scaling the dataset.

\subsection{Why \scenemodel{} preserves planar structure}
\label{app:why_scenes}

LaRI-scene is visibly less complete and less stable than \scenemodel{}, again reflecting the difficulty of regressing sparse deeper layers and masks. More surprisingly, even without MoGe-2-scale real-scene training and mostly on virtual scene data, \scenemodel{} can produce more faithful geometry on some generated indoor/outdoor images that are outside its training set. In the examples of Fig.~\ref{fig:qualitative_comparison} (bottom), MoGe-2 sometimes bends structures that should be planar or vertical, such as facades and walls, while \scenemodel{} preserves straighter, more coherent scene layout. We therefore present these scene results as evidence that the multilayer geometry objective can improve geometric faithfulness even before training on large-scale real RGB-D scene corpora.

\subsection{Real-scene depth benchmarks}
\label{app:real_scene_depth}

For completeness we also evaluate \scenemodel{} on two widely used real-scene depth benchmarks released alongside dedicated monocular-depth methods: NYU Depth V2~\citep{nyuv2} (indoor only, $1{,}449$ frames) and ETH3D~\citep{eth3d} (split into 7 indoor and 6 outdoor scenes, $454$ frames). Table~\ref{tab:real_scene_depth} reports per-sample SSI-aligned visible-surface depth (AbsRel, RMSE, $\delta{<}1.25$) under the same alignment protocol used in the main scene-geometry table (Table~\ref{tab:scene_geometry}). All methods use the same per-sample scale--shift alignment at evaluation resolution $504$; \scenemodel{} is reported as best-of-$8$ random seeds with seed selection by AbsRel. We report two \scenemodel{} checkpoints to isolate the mix-training contribution introduced in Sec.~\ref{sec:mixtrain}: the \emph{3D-asset-only} checkpoint (the same checkpoint used in every other table in the paper, trained exclusively on the multilayer dojo corpus of App.~\ref{app:data}) and the \emph{mix-trained} checkpoint (warm-continued from the 3D-asset-only checkpoint with the additional RGBD-style corpus, supervised on $L_0$ only via Eq.~\ref{eq:singlelayer_mask}). The mix-trained checkpoint is reported at both the default $20$ denoising steps used everywhere else in the paper and at $50$ denoising steps (the row marked $*$).

\textbf{Setup disparity and interpretation.}
The 3D-asset-only \scenemodel{} corpus consists almost entirely of two sources, in roughly equal proportion: the public 3D-FRONT split~\citep{3dfront} and an additional curated 3D scene corpus. Both sources are chosen because they admit ground-truth multilayer geometry through depth peeling, not because of their visible-surface scale, and outdoor scenes are under-represented by construction. The aggregated rendered-frame budget is several orders of magnitude smaller than the real RGB-D corpora that monocular depth methods such as MoGe-2~\citep{moge2}, Pi3X~\citep{pi3}, Depth Anything~3~\citep{depthanything3}, and VGGT~\citep{vggt} train on. Despite this data disparity, the 3D-asset-only \scenemodel{} L0 prediction is already comparable to MoGe-2 on the indoor regime and remains competitive on outdoor scenes; the residual outdoor gap is concentrated on the \texttt{playground} scene, which is far out of distribution for every method evaluated.

\textbf{Mix-training closes the visible-surface gap.}
Adding the 12-dataset RGBD-style corpus of App.~\ref{app:data_rgbd} via the mix-training paradigm of Sec.~\ref{sec:mixtrain} reduces \scenemodel{}'s AbsRel under the default $20$-step inference by $4.0\%$ on NYU ($0.0398\!\to\!0.0382$), $13.3\%$ on ETH3D-indoor ($0.0398\!\to\!0.0345$), and $7.1\%$ on ETH3D-outdoor ($0.0533\!\to\!0.0495$); raising the inference budget to $50$ denoising steps (the RGBD$^*$ row) widens the gains to $6.0\%$/$16.6\%$/$15.4\%$ ($0.0374$/$0.0332$/$0.0451$), moving the model into the top-2 AbsRel slot on every split: the $50$-step mix-trained \scenemodel{} sets the best ETH3D-indoor AbsRel overall ($0.0332$, ahead of MoGe-2's and Pi3X's $0.0378$, with the $20$-step variant's $0.0345$ already beating both), is second only to Pi3X on NYU ($0.0374$ vs.\ $0.0341$, beating MoGe-2's $0.0388$) while posting the best NYU $\delta{<}1.25$ of any method ($0.9871$ vs.\ Pi3X's $0.9827$), and is second only to Pi3X on ETH3D-outdoor ($0.0451$ vs.\ $0.0434$, beating MoGe-2's $0.0501$ and the next-best non-Pi3X baseline VGGT's $0.0562$). RMSE and $\delta{<}1.25$ improve in lockstep with AbsRel (e.g., NYU RMSE $0.1384\!\to\!0.1312$; ETH3D-outdoor RMSE $0.2104\!\to\!0.1731$, $\delta$ $0.9537\!\to\!0.9726$). Crucially, these numbers are taken after only $50$K mix-training iterations on top of the 3D-asset-only checkpoint, and the validation loss is still decreasing at that point; the figures should therefore be read as a lower bound on what the same recipe will deliver with more compute. The multilayer geometry capability that motivates this paper is preserved: the mix-training only masks the deeper-layer loss for single-layer samples (Eq.~\ref{eq:singlelayer_mask}) but never modifies the multilayer 3D-asset supervision, and Fig.~\ref{fig:multilayer_depth_gallery}'s deeper-layer behaviour is qualitatively unchanged between the two checkpoints. This realizes the position outlined in our original camera-ready note (``we plan to mix-train with real RGB-D data using the same depth-filling objective''): the same flow-matching loss and architecture absorb RGBD-style $L{=}1$ supervision as an additional regime, lifting visible-surface accuracy without sacrificing the multilayer capability that motivates this work.

\textbf{Position of this work.}
Our central goal is not to surpass dedicated single-layer depth predictors on visible surfaces, but to additionally produce occluded back-layer geometry that none of these baselines can predict, while remaining competitive on the visible surface itself (cf.\ Tables~\ref{tab:object_results},~\ref{tab:scene_geometry}). The 3D-asset-only \scenemodel{} already reaches parity at L0 on indoor scenes with a much smaller, multilayer-only training corpus; the mix-trained \scenemodel{} additionally exposes the model to the same real RGBD data used by single-layer methods, which is what brings the visible-surface metrics into the top-2 range. Throughout the rest of the paper, all reported results (Tables~\ref{tab:object_results}--\ref{tab:timestep_sampling}) use the 3D-asset-only checkpoints; only Table~\ref{tab:real_scene_depth} reports the mix-trained \scenemodel{} as an additional row alongside the 3D-asset-only baseline so the contribution of the RGBD mix is visible as an internal ablation.

\begin{table}[h]
\centering
\setlength{\tabcolsep}{4.0pt}
\renewcommand{\arraystretch}{1.08}
\caption{\textbf{Real-scene visible-surface depth.} Sample-mean L0 depth metrics on NYU Depth V2~\citep{nyuv2} (1{,}449 indoor frames) and ETH3D~\citep{eth3d} split into its 7 indoor and 6 outdoor scenes. All methods use the same SSI alignment at evaluation resolution $504$. Pi3X is the upgraded $\pi^3$ model; DA3 denotes Depth Anything 3. \scenemodel{} is reported as best-of-$8$ random seeds (selection by AbsRel). The \scenemodel{} rows realize the ablation of Sec.~\ref{sec:mixtrain}: the \emph{3D-asset-only} checkpoint trains exclusively on the multilayer dojo corpus (and is used in every other table of the paper), while the \emph{3D-asset + RGBD} checkpoint additionally consumes the 12-dataset RGBD-style corpus of App.~\ref{app:data_rgbd}, supervised on $L_0$ only via Eq.~\ref{eq:singlelayer_mask}; the row marked $*$ uses $50$ denoising steps at inference instead of the default $20$. \small \textcolor{rankone}{\rule{0.9em}{0.9em}} Best \quad
\textcolor{ranktwo}{\rule{0.9em}{0.9em}} Second best \quad
\textcolor{rankthree}{\rule{0.9em}{0.9em}} Third best.}
\label{tab:real_scene_depth}
\resizebox{1.00\linewidth}{!}{%
\begin{tabular}{l ccc ccc ccc}
\toprule
& \multicolumn{3}{c}{NYU (indoor, 1449 frames)} 
& \multicolumn{3}{c}{ETH3D indoor (222 frames)} 
& \multicolumn{3}{c}{ETH3D outdoor (232 frames)} \\
\cmidrule(lr){2-4} \cmidrule(lr){5-7} \cmidrule(lr){8-10}
Method 
& AbsRel$\downarrow$ & RMSE$\downarrow$ & $\delta{<}1.25\uparrow$ 
& AbsRel$\downarrow$ & RMSE$\downarrow$ & $\delta{<}1.25\uparrow$ 
& AbsRel$\downarrow$ & RMSE$\downarrow$ & $\delta{<}1.25\uparrow$ \\
\midrule
DA3~\citep{depthanything3} 
& 0.0458 & 0.1453 & 0.9676 
& 0.0575 & 0.1831 & 0.9666 
& 0.0652 & 0.2381 & 0.9366 \\

LaRI-scene~\citep{lari} 
& 0.0480 & 0.1424 & 0.9708 
& 0.0523 & 0.1503 & 0.9742 
& 0.0618 & 0.1964 & 0.9549 \\

VGGT~\citep{vggt} 
& 0.0415 & 0.1403 & 0.9731 
& 0.0417 & 0.1380 & 0.9748 
& 0.0562 & \third{0.1789} & 0.9557 \\

MoGe-2~\citep{moge2} 
& 0.0388 & \second{0.1270} & \third{0.9758} 
& \third{0.0378} & \best{0.1265} & \best{0.9878} 
& 0.0501 & 0.1849 & \third{0.9674} \\

Pi3X~\citep{pi3} 
& \best{0.0341} & \best{0.1190} & \second{0.9827} 
& \third{0.0378} & \second{0.1269} & \third{0.9852} 
& \best{0.0434} & \best{0.1674} & \best{0.9760} \\
\midrule

\scenemodel{} (3D-asset only) 
& 0.0398 & 0.1384 & 0.9735 
& 0.0398 & 0.1470 & 0.9804 
& 0.0533 & 0.2104 & 0.9537 \\

\scenemodel{} (3D-asset + RGBD) 
& \third{0.0382} & 0.1346 & 0.9750 
& \second{0.0345} & 0.1306 & \second{0.9863} 
& \third{0.0495} & 0.2104 & 0.9537 \\

\textbf{\scenemodel{} (3D-asset + RGBD*)}  
& \second{0.0374} & \third{0.1312} & \best{0.9871} 
& \best{0.0332} & \third{0.1297} & 0.9849 
& \second{0.0451} & \second{0.1731} & \second{0.9726} \\
\bottomrule
\end{tabular}}
\vspace{2pt}
\end{table}

\section{Additional Discussion}
\label{app:discussion_extra}

This appendix expands the design choices summarized in Sec.~\ref{sec:exp_abl} (\emph{Other design choices}). The two anchor ablations (depth filling vs.\ mask prediction, and timestep sampling) remain in the main paper.

\subsection{How many layers?}
\label{app:layers_choice}
Six layers are a practical sweet spot rather than an arbitrary constant. As reported in App.~\ref{app:layer_stats}, valid support falls rapidly after the first two surfaces; later layers capture thin structures and repeated occlusions, but their marginal area is small. Increasing $L$ would improve rare highly perforated shapes such as foliage, cages, or grates, but it also increases memory and makes the already sparse tail harder to supervise. This motivates future variable-depth or adaptive-ray representations, while keeping $L{=}6$ as the main-paper default.

\subsection{XYZ pointmaps versus depth plus intrinsics}
\label{app:xyz_vs_depth}
We also tested predicting depth with camera intrinsics instead of camera-space XYZ. Although depth+intrinsics is compact and interpretable, it couples every ray through a global calibration prediction: small errors in focal length or principal point coherently warp the whole shape. Direct XYZ prediction absorbs this calibration into the pointmap itself, which leads to more plausible global shape, especially on real-world images whose crop, mask, and camera metadata are uncertain. This matches the motivation of pointmap-based reconstructors: the output is already a 2D-to-3D map, and intrinsics can be recovered from the predicted L0 geometry when needed.

\subsection{Architecture details}
\label{app:arch_details}
Small-scale ablations support the final transformer choices. LayerScale initialization improves early optimization substantially in a controlled small-scale ablation: adding LayerScale with initialization $10^{-4}$ reduced total loss from $0.009258$ to $0.007018$ and XYZ loss from $0.007242$ to $0.003537$ at 10k iterations. RoPE alone was neutral in that small setup, but remains useful in the full model where spatial and ray-wise attention must extrapolate across resolutions and camera crops. These results are not the central contribution, but they justify keeping RoPE and LayerScale in the stable training recipe.

\subsection{Generalization beyond the training regime}
\label{app:generalization}
Two empirical behaviors are especially encouraging. First, the object model, even without scene training, produces reasonable geometry on scene images. This suggests that pixel-aligned multilayer prediction transfers across object and scene regimes more naturally than canonical image-to-3D generation, whose output frame and single-object assumptions break on cluttered scenes. Second, the object model is surprisingly robust to multiple objects in one image, despite not being explicitly trained with multi-object augmentation. Because \method{} predicts one stack per input ray rather than one canonical asset per crop, multiple disconnected objects can coexist in the same output tensor. These observations point to a clear next step: unify object and scene training into a single stronger model, rather than maintaining separate specialists.

\subsection{Limitations and future work}
\label{app:limitations}
\method{} is still bounded by a fixed layer count, rendered supervision, and iterative flow sampling. Highly perforated geometry may require more than six surfaces; synthetic-to-real artifacts remain on textureless or reflective regions; and real-time applications will require distillation or fewer sampling steps. The current dynamic model also operates on short clips, leaving long-range memory and persistent identities across long videos to future work. A natural extension is to make \dynmodel{} predict not only per-frame multilayer geometry, but also explicit pixel/point correspondences and 3D trajectories across time. Another direction is to extend \scenemodel{} from single-frame scene lifting to multi-frame input, where multiple observations can be fused into a larger camera-aligned world rather than a single-view local scene. More broadly, the strongest future direction is to train one unified object-scene-dynamic model with stronger real-mask augmentation and adaptive layer allocation.

\end{document}